\documentclass[sigconf]{acmart}

\usepackage{subcaption}
\AtBeginDocument{%
  \providecommand\BibTeX{{%
    \normalfont B\kern-0.5em{\scshape i\kern-0.25em b}\kern-0.8em\TeX}}}

\copyrightyear{2023}
\acmYear{2023}
\setcopyright{acmlicensed}\acmConference[GECCO '23]{Genetic and
Evolutionary Computation Conference}{July 15--19, 2023}{Lisbon, Portugal}
\acmBooktitle{Genetic and Evolutionary Computation Conference (GECCO '23),
July 15--19, 2023, Lisbon, Portugal}
\acmPrice{15.00}
\acmDOI{10.1145/3583131.3590460}
\acmISBN{979-8-4007-0119-1/23/07}

\begin{document}

\title{Learning to Act through Evolution of Neural Diversity in Random Neural Networks}


\author{Joachim Winther Pedersen}
\email{jwin@itu.dk}
\affiliation{%
  \institution{IT University of Copenhagen}
  \streetaddress{2300 Copenhagen, Denmark}
  \city{Copenhagen}
  \country{Denmark}
  \postcode{43017-6221}
}

\author{Sebastian Risi}
\email{sebr@itu.dk}
\affiliation{%
  \institution{IT University of Copenhagen}
  \streetaddress{}
  \city{Copenhagen}
  \country{Denmark}}
\email{}


\begin{abstract}


Biological nervous systems consist of networks of diverse, sophisticated information processors in the form of neurons of different classes. In most artificial neural networks (ANNs), neural computation is abstracted to an activation function that is usually shared between all neurons within a layer or even the whole network; training of ANNs focuses  on synaptic optimization. In this paper, we propose the optimization of neuro-centric parameters to attain a set of diverse neurons  that can perform complex computations. Demonstrating the promise of the approach, we show that evolving neural parameters alone allows agents to solve various reinforcement learning tasks
\emph{without optimizing any synaptic weights}. 
While not aiming to be an accurate biological model, parameterizing neurons to a larger degree than the current common practice, allows us to ask questions about the computational abilities afforded by neural diversity in random neural networks.
The presented results open up interesting future research  directions,  such as combining evolved neural diversity 
 with activity-dependent plasticity.
\end{abstract}



\maketitle

\begin{figure}

\begin{center}
\includegraphics[scale=.28]{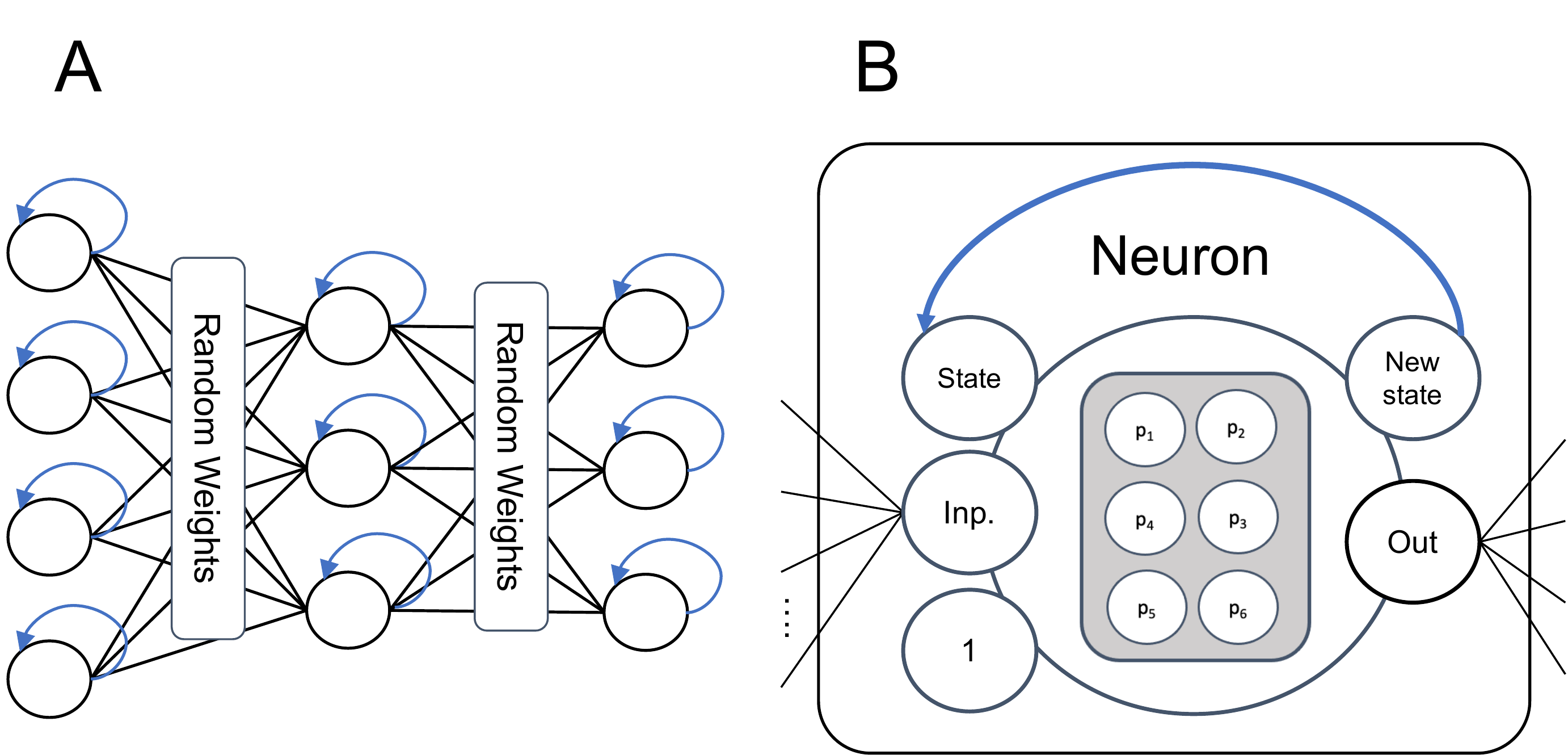}
\end{center}
\caption{Illustration of the proposed neural unit. \normalfont (A) Neural network  with random weights and a layer of neural units. (B) Zoomed-in view of a neural unit. The parameters $p_i$ are optimized in order to achieve an expressive function. These parameters are used to integrate the input with a neural state and a bias term through a vector-matrix multiplication. It outputs a value to be propagated to the next layer, as well as its own new state. For further details, see Section \ref{approach}. }
\centering
\label{fig:neuron_model}
\end{figure}

\section{Introduction}
\label{section:intro}
Brains of animals are characterized by the presence of neurons of many different classes \citep{lillien1997neural, soltesz2006diversity}. Each neural class has been shaped by evolution to contain unique properties that allow processing  incoming signals in different manners. Further, biological neurons are dynamical systems, capable of integrating information over time and responding to inputs in a history-dependent manner \citep{sekirnjak2002intrinsic, izhikevich2003simple, izhikevich2007dynamical, wasmuht2018intrinsic}.
Whereas the evolutionary optimization of biological neural networks has resulted in networks having multiple classes of recurrent processors, artificial neural networks (ANNs) tend to contain homogeneous activation functions, and optimization is most often focused on the tuning of synaptic weights. 
During the past decade, ANNs have become a dominating force within machine learning research \citep{schmidhuber2015deep, lecun2015deep, chauhan2018review, tay2022efficient}. As the name suggests, this technique originally took inspiration from biological networks found in the brains of animals \cite{fasel2003introduction, hassabis2017neuroscience}. The main analogous feature of ANNs to that of biological ones is that nodes in the network distribute information to each other and that the network learns to represent information through the gradual tuning of their interconnections. In ANNs, neuro-centric computation is abstracted to an activation function that is usually shared between all neurons within a layer or even the whole network.

Information in neural networks spreads through many synapses and converges at neurons. Given that a single biological neuron is a sophisticated information processor in its own right \citep{marder1996memory, beaulieu2018enhanced, beniaguev2021single}, it might be useful to reconsider the extreme abstraction of a neuron as being represented by a single scalar, and an activation that is shared with all other neurons in the network.
By parameterizing neurons to a larger degree than current common practice, we might begin to approach some of the properties that biological neurons are characterized by as information processors.

Inspired by the neural diversity found in biological brains, we introduce a parameterized neural unit with a feedback mechanism (Figure~\ref{fig:neuron_model}). The idea is that evolving a unique set of parameters for each neuron has the potential for creating a diverse set of neurons. 

While we are ultimately interested in potential synergies from evolving both weights and neural parameters, in order to investigate the expressive power of the proposed neural units in isolation, here we only evolve the parameters of neurons; the  randomly initialized synaptic weights are fixed throughout the entire process.

Our results show that even when never optimizing any synaptic parameters, the random networks with evolved neural units achieve performances competitive with simple baselines in different reinforcement learning tasks. For baseline comparisons, we evolve the weights of feedforward neural networks using a standard hyperbolic tangent function as non-linearities. We compare our approach both with (1) a small network with a similar number of weights as there are trainable parameters in the neural units of the main experiments, and (2) a network with the same number of neurons as in the main experiments and thus many more tunable synaptic weights. We also test a non-recurrent version of the neural units. The environments used are a variation of the classic CartPole environment \cite{gaier2019weight}, the \texttt{BipedalWalker-v3} environment, and the \texttt{CarRacing-v0} environment \citep{brockman2016openai}. In all cases, networks with evolved neurons performed on par with the weight-optimized networks.

By showcasing the potential of these more expressive and diverse neurons in fixed random networks, we hope to pave the way for future studies exploring potential synergies in the optimization of synaptic and neural parameters. More specifically, we believe that more expressive neural units like the ones proposed here could be useful in combination with online synaptic activity-dependent plasticity functions \citep{van1994activity, abbott2000synaptic, stiles2000neural, soltoggio2018born, najarro2020}.

\section{Related Work}
\label{related}

\textbf{Neurocentric Optimization.} Biases in neural networks are examples of neuro-centric parameters. When an ANN is optimized to solve any task, the values of the network's weights and biases are gradually tuned until a functional network has been achieved. The network most commonly has a one bias value for each neuron and the weight parameters thus greatly outnumber these neuro-centric bias parameters. It is well known that the function of biases is to translate the activations in the network \citep{benitez1997artificial} and ease the optimization of the network. 

Another well-known example of neuro-centric parameters is found in the PReLU  activation functions \citep{he2015delving} where a parameter is learned to determine the slope of the function in the case of negative inputs. Introducing this per-neuron customization of the activation functions was shown to improve the performance of networks with little extra computational cost.

Neuro-centric parameter optimization can also be found within the field of plastic neural networks. In one of their experiments,  Urzelai and Floreano~\citep{urzelai2001evolution} optimized plasticity rules for each neuron (they referred to this as 'node encoding'), such that each incoming synapse to a node was adapted by a common plasticity rule.
The idea of neuro-centric parameters is thus far from new. However, in contrast to earlier work, in this paper, we explore the potential of solely optimizing neuro-centric parameters in a randomly initialized network without ever adapting the weights.  

\textbf{Activation Functions in Neuroevolution.} How artificial neurons are activated has a major impact on the performance of ANNs \cite{nwankpa2018activation}. Evolution has been used to discover activation functions to optimize performance in networks that are optimized on supervised classification problems using backpropagation \citep{basirat2018quest, liu2020evolving,  bingham2020evolutionary, bingham2022discovering}. These approaches aim to find a single optimal activation of neurons that all neurons within a layer or a whole network share.

Not all ANNs have a single activation function for all hidden neurons. Some versions of Neuro-Evolution of Augmented Topologies (NEAT) \citep{stanley2002evolving, papavasileiou2021systematic} allow for different activation functions on each neuron. The NEAT algorithm searches through networks with increasing complexity over the process of evolution. Starting from a simple network structure, each new network has a chance of adding a new neuron to the network. When a new neuron is added, it can be allocated a random activation function from a number of predetermined functions. In newer versions of NEAT, mutations allow activation functions of neurons to be changed even after it was initially added \citep{hagg2017evolving}. This resulted in more parsimonious networks.

In their \emph{weight agnostic neural network} (WANN) work, \citeauthor{gaier2019weight}~\cite{gaier2019weight} used NEAT to find network structures that could perform well, even when all weights had the same value. Notably, hidden neurons could be evolved to have different activation functions from each other. This likely extended the expressiveness of the WANNs considerably.
Our work is similar in spirit to that of Gaier and Ha, in that we are also exploring the capabilities of a component of neural networks in the absence of traditional weight optimization. However, in our work, all networks have a standard fully connected structure. Furthermore, we do not choose from a set of standard activation functions, but introduce stateful neurons with several parameters to tune for each neuron.

\textbf{Lottery Tickets \& Supermasks.} The Lottery Ticket Hypothesis \citep{frankle2018lottery, frankle2019stabilizing} states that the reason large ANNs are trainable is that a large network contains a combinatorial number of subnetworks, one of which is likely to be easily trainable for the task at hand. The hypothesis is based on the finding that after having trained a large network, it is usually possible to prune a large portion of the parameters without suffering a significant loss in performance \citep{li2016pruning}.
An even stronger take on the Lottery Ticket Hypothesis states that due to the sheer number of subnetworks that are present within a large network, it is possible to learn a binary mask on top of the weight matrices of a randomly initialized neural network, and in this manner get a network that can solve the task at hand \citep{malach2020proving, wortsman2020supermasks, ramanujan2020s}. This has even been shown to be possible at the level of neurons; with a large enough network initialization, a network can be optimized simply by masking out a portion of the neurons in the network \citep{wortsman2020supermasks, malach2020proving}.

Learning parameterized versions of neurons could be seen as learning a sophisticated mask on top of each neuron as opposed to a simple binary mask. The idea of learning a binary mask on top of a random network relies on the random network being sufficiently large \citep{malach2020proving} so that the chance of it containing a useful subnetwork is high. Masking neurons is a less expressive masking method than masking weights: it is equivalent to masking full columns of the weight matrices instead of strategically singling out weights in the weight matrix. As such, the method of masking neurons with binary masks requires larger random networks to be successful.
In this paper, we optimize neuro-centric parameters in relatively small networks. This is possible because the neurons themselves are much more expressive than a binary mask.

\section{Evolving Diverse Neurons in Random Neural Networks}

\label{approach}
Typically, optimization of ANNs has been framed as the learning of distributed representations \citep{bengio2013representation} that can become progressively more abstract with the depth of the network. Optimization of weights, is a process of fine-tuning the iterative transformation of one representation into another to end up with a new, more useful representation of the input.
How the intermediate layers respond to a given input depends on the specific configuration of the weight matrix responsible for transforming the input as well as their activation function. 

Randomly-initialized networks can already perform useful computations \citep{ulyanov2018deep, he2016powerful, hochreiter1997long}. 
When neural units are trained specifically to interpret signals from a fixed random matrix, as is the case in this paper, the initially arbitrary transformations will become meaningful to the function, as long as there exist detectable patterns between the input the function receives and the output that the function passes on and is evaluated on. Whether a pattern is detectable depends on the expressiveness of the function. 
With this in mind, it is reasonable to assume that if neurons in the network are made more expressive, they can result in useful representations even when provided with arbitrary transformations of the input.

Motivated by the diversity of neuron types in animal brains \citep{soltesz2006diversity}, we aim to test how well a neural network-based agent can perform  reinforcement learning tasks through optimization of its neuro-centric parameters alone without optimizing any of its neural network weights. 
An illustration of the neural model we optimize in this paper is shown in Fig.~\ref{fig:neuron_model}. Each neural unit consists of a small three-by-two matrix of values to be optimized.
Each neural unit in a layer is at each time step presented with a vector with three elements. The input value, propagated through the random connection from the previous layer, is concatenated with the current state of the neuron and a bias term. Together, these form a vector. The output of a neuron is the vector-matrix multiplication. From the perspective of a single neuron, this can be written as:
\begin{equation}
 [\hat{x}_{l,i}^t, h_{l,i}^t] = tanh(\mathbf{n}_{l,i} \cdot [ x_{l,i}^t, h_{l,i}^{t-1}, 1]^T ).
\end{equation}

Here, $x_{l,i}^t$ is the input value, $h_{l,i}^{t-1}$ is the state of the neuron, and $\mathbf{n_{l,i}}$ the matrix of neural parameters, with $l$ denoting the current layer in the network, $i$ the  placement of the neuron in the layer, and $t$ is the current time step. The hyperbolic tangent function is used for non-linearity and to restrict output values to be in [-1, 1]. $\hat{x}_{l+1,j}^t$ is the value that is propagated through weights connecting to the subsequent layer, and $h_{l,i}^t$ is the updated state of the neuron.
As the three-by-two matrix has six values in total, we need to optimize six parameters for each neuron in the network. 

Representing a neuron by a small matrix means that the neuron can take more than a single value as input, as well as output more than one value. 
Here, we utilize this to endow each neuron with a state. The state of the neuron is integrated with the input through the optimized neural parameters. Part of the neuron's output becomes the new state of the neuron, which is fed back to the neuron with the next input. This turns our neurons into small dynamical systems. Presented with the same input value at different points in the neuron's history can thus yield different outputs. We find that stateful neurons provide a convenient and parameter-efficient way of equipping a network with some memory capabilities. One can see a layer of such neurons as a set of tiny recurrent neural networks (RNNs) that are  updated in parallel with local inputs, unique to each RNN. As such, a layer of this proposed neural unit differs from simple RNN architectures, such as Jordan Networks \citep{jordan1997serial} or Elman Networks \citep{elman1990finding} in that a state associated with a neuron only affects the next state and output of that particular neuron. These local recurrent states only rely on the small matrix of the neural unit, i.e., $n$ times six parameters, where $n$ is the number of neurons in the layer. A recurrent layer of, e.g, an Elman Network requires an $n$-by-$n$ sized matrix to feed its activations back into itself. Additionally, the calculation of the neural state and the output of the neuron are separated to a higher degree for the neural units, compared to the hidden state being a copy of the neural output.

\section{Experiments}
\label{experiments}

We optimize neural units in otherwise standard fully-connected feedforward neural networks. All networks in our experiments have two hidden layers, containing $128$ and $64$ neurons, respectively. We use learned neural units for all neurons, including in the input and output layers. The sizes of the input and output layers vary with the environments described below. The fixed weight values are sampled from $ \mathcal{N}(0,0.5)$. We ran each experiment three times on different seeds, except for the weight-optimized models in the Car Racing environment, which we ran only twice due to its longer training times.

While neurons with recurrent states are common in the field of spiking neural networks that emphasizes biological realism \cite{gerstner1990associative,izhikevich2003simple, izhikevich2007dynamical, garaffa2021revealing}, it is a departure from the simple neurons found in most ANNs. As a control, we also optimize neuro-centric parameters for neurons without a recurrent state (\textbf{Simple Neuron}). The setup for optimizing these is very similar to that of the stateful neurons, but no part of the output of the vector-matrix multiplication is fed back to the neuron's input at the next time step. Instead of being represented by three-by-two matrices, these simple neurons are represented by two-by-one matrices (vectors in this case) and thus have fewer parameters to optimize. As such, the parameters of these simple neurons are simply a scalar of the neural input and a bias.

As baselines, we  optimize weights of standard feedforward networks. For these, we run two different settings: one has a similar number of adjustable parameters as the number of parameters in the neural unit approach (\textbf{Small FFNN}). To get the number of weights to be similar to neural parameters, the widths and depths of these networks have to be considerably smaller than the random networks used in the main experiments. In the second baseline setting (\textbf{Same FFNN}), we train weights of networks that have the same widths and depths as the networks in the main experiments, and thus many more adjustable parameters. Unless stated otherwise, the activation function for all baseline experiments is the hyperbolic tangent function for all neurons.

\subsection{Environments}

We test the effectiveness of evolving a diverse set of neurons in randomly initialized networks in three diverse continuous control tasks: the CartPoleSwingUp environment \citep{gaier2019weight}, the Bipedal Walker environment, and the Car Racing environment \citep{brockman2016openai}, which are described below: 

\begin{figure}

\includegraphics[scale=.135]{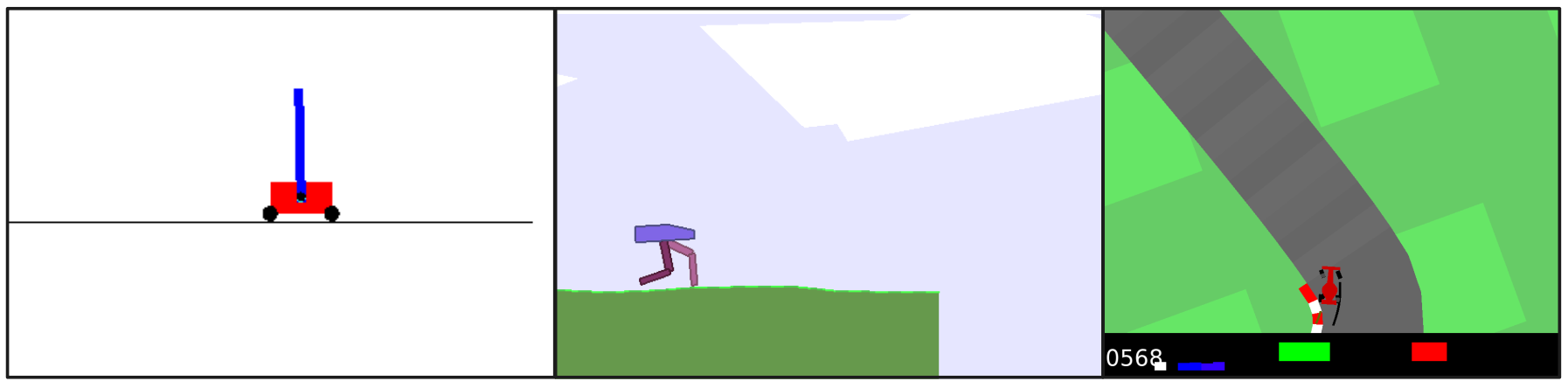}

\caption{Environments used for experiments. \normalfont Left:  \texttt{CartPoleSwingUp}. Middle: \texttt{BipedalWalker-v3}. Right: \texttt{CarRacing-v0}. } 
\centering
\label{envs_im}
\end{figure}

\textbf{CartPoleSwingUp.}
This environment is a variation of the classic control task \citep{barto1983neuronlike}, where a cart is rewarded for balancing a pole for as long as possible (Fig.~\ref{envs_im}; left). In the \texttt{CartPoleSwingUp} variation, an episode starts with the pole hanging downwards, and to score points, the agent must move the cart such that the pole gets to an upright position. From there, the task is to keep balancing the pole. We use the implementation of Gaier and Ha \citep{gaier2019weight}. The agent gets five values as input, and must output a single continuous value in $[-1,1]$ in order to move the cart left and right. With these input- and output layers, the total number of neurons in the network becomes $198$, and we optimize 1,259 parameters. For a feedforward network with a similar number of weights, we optimize a network with two hidden layers, both of size $32$.

\textbf{Bipedal Walker.} To get a maximum score in the \texttt{BipedalWalk \newline er-v3} environment \citep{brockman2016openai}, a two-legged robot needs to learn to walk as efficiently and robustly as possible (Fig.~\ref{envs_im}; middle). The terrain is procedurally generated with small bumps that can cause the robot to trip if its gait is too brittle. Falling over results in a large penalty to the overall score of the episode. Observations in this environment consist of $24$ values, including LIDAR detectors and information about the robot's joint positions and speed. For actions, four continuous values in $[-1,1]$ are needed. This result in a network with $220$ neurons altogether.
To get a network with a similar number of weights as there are parameters in the main experiment, we train a feedforward network with two hidden layers of size $32$ and $16$, respectively.

\textbf{Car Racing.} 
In the \texttt{CarRacing-v0} domain \citep{brockman2016openai}, a car must learn to navigate through procedurally generated tracks (Fig.~\ref{envs_im}; right). The car is rewarded for getting as far as possible while keeping inside the track at all times.
To control the car, actions consisting of three continuous values are needed. One of these is in $[-1,1]$ (for steering left and right), while the other two are in $[0,1]$, one for controlling the gas, and the other for controlling the break.
The input is a top-down image of the car and its surroundings. For the experiments in this paper, we wanted to focus on the effectiveness of the neural units in fully connected networks. We, therefore, followed a strategy closely mimicking that used by \citeauthor{najarro2020}  \cite{najarro2020} to get a flat representation of the input image. We normalize the input values and resize the image into the shape of $3x84x84$. The image is then sent through two convolutional layers, with the hyperparameters of these specified in Table~\ref{tab:cnn}. After both layers, a two-dimensional max pooling with kernel size $2$ and stride $2$ was used to gradually reduce the number of pixels.

\begin{table}
\begin{center}
  \caption{Convolutional Layer Parameters.}

  \label{tab:cnn}
  
  \begin{tabular}{ccccccc}
   \toprule
     \hline
      &Layer 1 & Layer 2 & \\
    \midrule
       Input Channels & 3              & 6  \\
       Output Channels & 6              & 8  \\
       Kernel Size  &   3           & 5 \\ 
      Stride  & 1 & 2 \\
      Activation Function & tanh                 & tanh  \\
     Bias &        Not used      & Not used \\  \hline
   \bottomrule

\end{tabular}
\end{center}
\end{table}

The output of the convolutional layers is flattened, resulting in a vector containing $648$ values. This vector is then used as input to a fully connected feedforward network with our proposed neural units. Importantly, the parameters of the convolutional layers stay fixed after initialization and are never optimized.
The output layer has three neurons. With the much larger input layer, this network has $844$ neurons and 5,218 adjustable parameters. Since two of the action values should be in $[0,1]$, a sigmoid function is used for these two output neurons in place of the hyperbolic tangent function as shown in Equation 1.

For the baseline experiments, we use the same strategy of  convolutional layers with fixed parameters to get a flat input for the feedforward networks, the weights of which we are optimizing. Since the input is so large, the feedforward network can only have a single hidden layer of size $8$ to get a similar number of adjustable parameters as in the main approach. This results in a network with 5,219 parameters, including weights and biases. The activation function for all neurons in the network is the hyperbolic tangent function, except for two of the output neurons, which are activated by the sigmoid function.

\subsection{Optimization Details}
\label{optimization}

For parameter optimization, we use a combination of a Genetic Algorithm (GA) and Covariance Matrix Adapatation Evolution Strategy (CMA-ES) \cite{hansen2006cma}. More specifically, GA is used for the first 100 generations of the optimization. 
The GA used here searches the parameter space broadly with a larger population size than CMA-ES. The best solution found by the GA is then used as a good starting point for the CMA-ES algorithm to continue evolution. Starting with a GA search with a large sigma, was in preliminary experiments found to help the CMA-ES avoid getting stuck early at a local optimum.
For both algorithms, we use off-the-shelf implementations provided by \citeauthor{ha2017evolving}~\cite{ha2017evolving} and \citeauthor{hansen2006cma}~\cite{hansen2006cma}.
For all experiments, the GA uses a population size of $512$, and its mutations are drawn from a normal distribution $ \mathcal{N}(0,1)$. All other hyperparameters are the default parameters of the implementation. The large sigma means that the GA can cover a large area but in a coarse manner. For the CMA-ES algorithm, we use a population size of $128$, and set weight decay to zero. Other than that, hyperparameters are the default parameters of the implementation.
The total number of generations is 1,400 for the \texttt{Car Racing} environment, and 4,000 for the \texttt{BipedalWalker} and \texttt{CartPoleSwingUp} environments.

For the large weight-optimized networks, CMA-ES becomes impractical to use due to its use of the covariance matrix of size $N^2$ with $N$ being the number of parameters to optimize. For this reason, we instead use Evolution Strategy (sometimes referred to as ``OpenES''  \citep{ha2017visual}) as described by \citeauthor{salimans2017evolution} \cite{salimans2017evolution} and. We use a population size of $128$ and otherwise use the default parameters of the implementation \citep{ha2017visual}.

\section{Results}

\begin{figure*}
\begin{center}
\includegraphics[scale=.3]{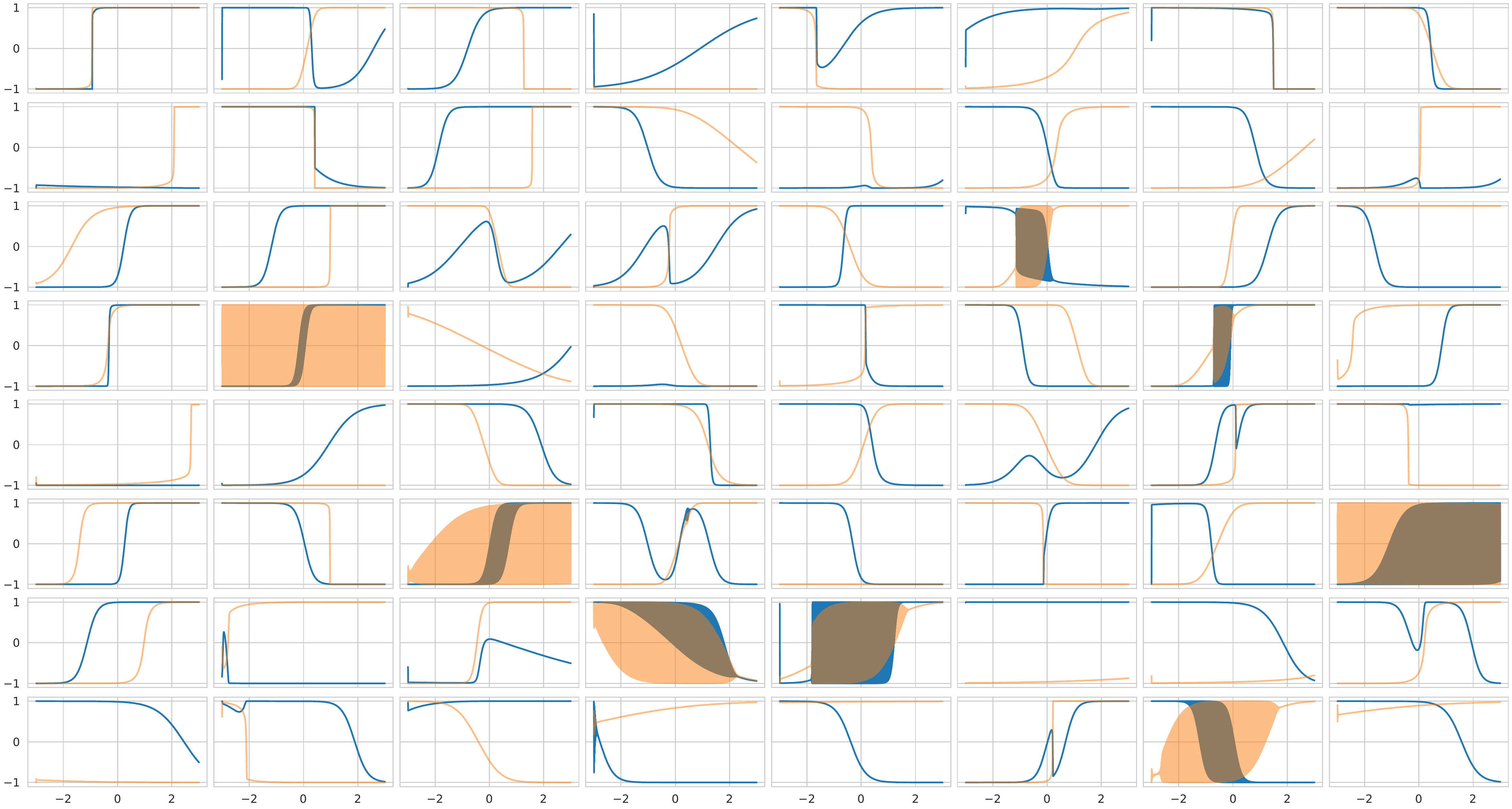}
\end{center}
\caption{Evolved Neural Diversity. \normalfont Displayed are each of the 64 activated neurons of the second hidden layer optimized to solve the \texttt{CarRacing} task. In blue are the activations that are passed on to the next layer. In orange are the neural states. One thousand inputs are given from $-3$ to $3$ in an ordered manner, from most negative to most positive. Given the updatable neural state, the ordering of the inputs matter, and a different ordering would have yielded different plots. For all plots, the neural state is initialized as zero before the first input. Several of the found activations look like we would expect the hyperbolic tangent function with a bias and/or the possibility of a negative sign to look. Few functions seem unresponsive to the input. However, many functions are clearly both responsive and different from the standard form of the hyperbolic tangent function. We find functions that have oscillatory behavior in some or all of the input space. This  is where the graphs of outputs and/or neural states color areas of the graph due to rapid changes with the input. Other functions are non-monotonic and have peaks and valleys in particular areas. }
\centering
\label{fig:functions}
\end{figure*}

\begin{figure*}
\begin{center}
\includegraphics[scale=.3]{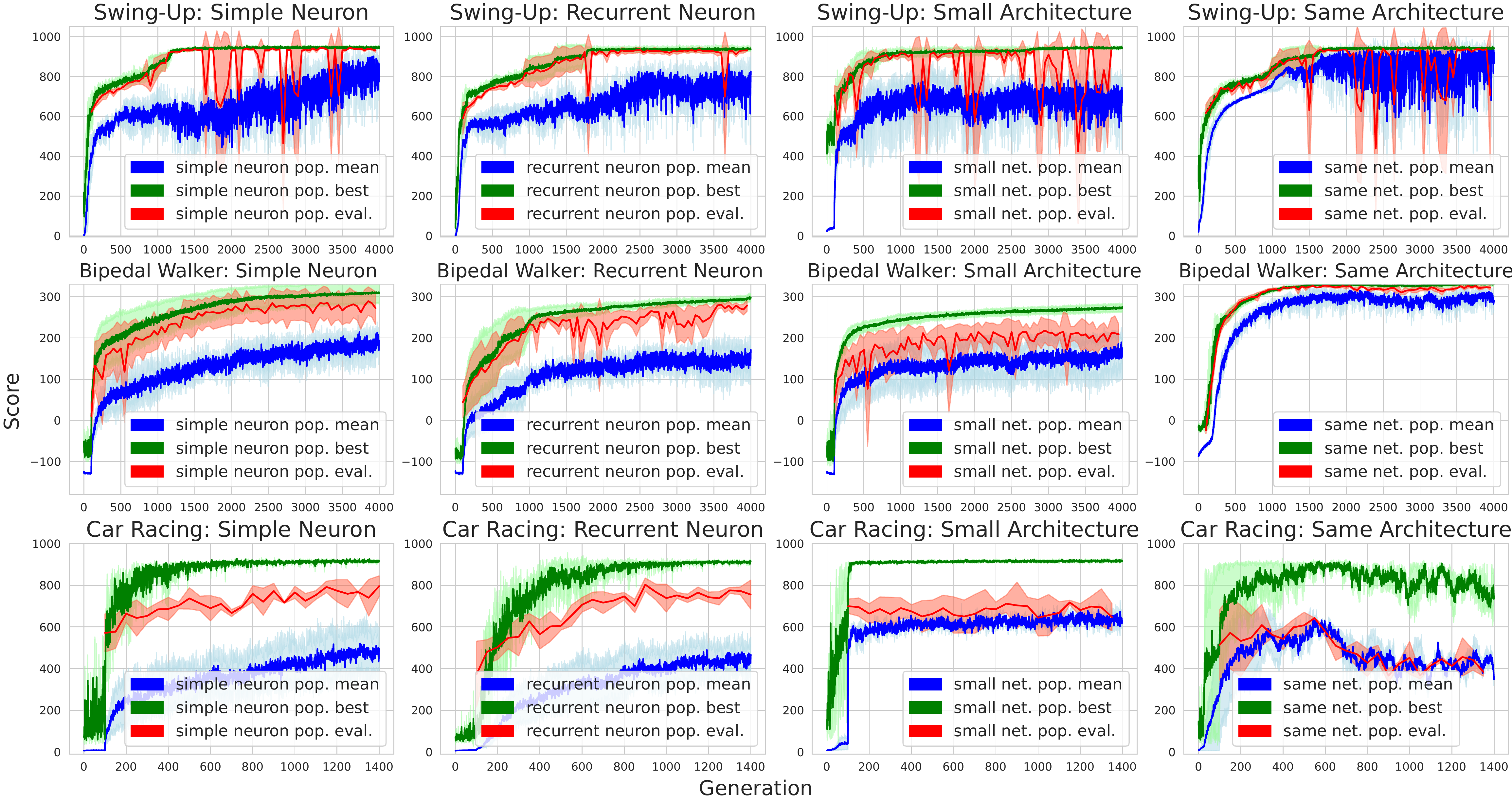}
\end{center}
\caption{Training curves for all experimental settings. \normalfont Means and standard deviations for populations of runs on different seeds. Each setting was run three times except for the weight-optimized models in the Car Racing environment, which were run twice. Every $50^{th}$ generation, the current solution was evaluated $64$ times (red).}
\centering
\label{fig:curves}
\end{figure*}

\begin{figure*}
\begin{center}
\includegraphics[scale=.3]{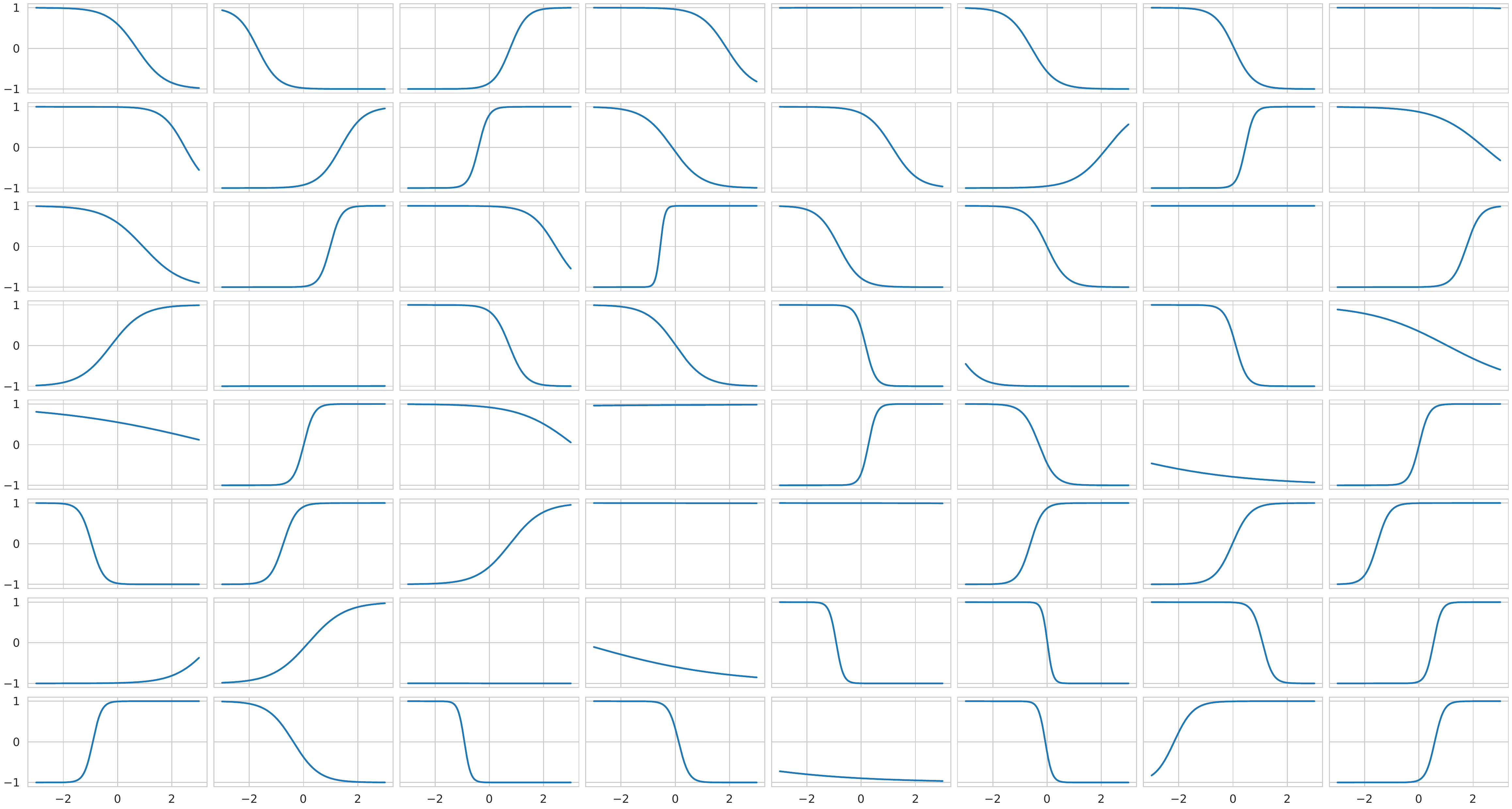}
\end{center}
\caption{Activations of Non-Recurrent Neurons. \normalfont Displayed are each of the 64 activations of the second hidden layer optimized to solve the \texttt{CarRacing} task. One thousand inputs are given from -3 to 3 in an ordered manner, from most negative to most positive. Neural activations are either monotonically increasing or decreasing, or unresponsive to the input.}
\centering
\label{fig:simple_functions}
\end{figure*}

Evaluations of the most successful runs of each experimental setting are summarized in Table \ref{tab:eval}. All experimental settings achieved good scores on the \texttt{CartPoleSwingUp} task. Only the weight-optimized network with hidden layers of size $128$ and $64$ managed to average a score above $300$ points over $100$ episodes in the \texttt{BipedalWalker-v3} environment. However, the optimized recurrent neurons in a random network came close with just a fraction of the number of optimized parameters. The smaller weight-optimized network (\textbf{Small FFNN}) and the simple neurons achieved similar scores of $240$ and $239$, respectively. This is indicative of the agent having learned to walk to the end of the level in most cases but in an inefficient manner.

In the \texttt{CarRacing-v3} environment, the agent based on recurrent neurons scored the highest, though none of the approaches reached an average score above $900$ over $100$ episode, which is needed for the task to be considered solved. However, with a mean score above 800, the agent was able to successfully complete the majority of the procedurally generated test episodes. Training curves for all experimental settings can be found in Figure \ref{fig:curves}. 

\label{results}
\begin{table}
\label{results_table}
\begin{center}
  \caption{Table of Results. \normalfont Means and standard deviations over $100$ episodes, and number of parameters optimized for each experimental setting. 
  Scores are evaluated with the most successful run in each setting. For context, results from Weight Agnostic Neural Networks (WANN) \citep{gaier2019weight} are included as another method that does not optimize weights. Number of parameters listed for WANN are the final number of connections in the evolved structure. }
  \label{tab:eval}
    
  \begin{tabular}{ccccccc}
   \toprule
     \hline
     & Model & Score & \# Param.\\
\hline
    \midrule
     & Simple Neurons      & 892  $\pm$ 177 & 396 \\
     & Rec. Neurons      & 916  $\pm$ 79 & 1,259 \\
\texttt{CartPoleSwingUp} & Small FFNN   & 805 $\pm$ 296 & 1,281  \\
     & Same FFNN    & \textbf{922}  $\pm$ 73 & 9,089 \\  
     & WANN    & 732  $\pm$ 16 & 52 \\       
     \hline     
    \midrule
     & Simple Neurons      & 239  $\pm$ 52 & 440 \\
     & Rec. Neurons      & 295 $\pm$ 63  & 1,325  \\
\texttt{BipedalWalker} & Small FFNN   & 240 $\pm$ 59 & 1,396  \\
     & Same FFNN    & \textbf{318} $\pm$ 46 & 11,716 \\
     & WANN    & 261 $\pm$ 58 & 210 \\ 
     \hline
    \midrule
     & Simple Neurons      & 820  $\pm$ 118 & 1,686 \\
     & Rec. Neurons      & \textbf{822} $\pm$ 74  & 5,218  \\
\texttt{CarRacing} & Small FFNN    & 798 $\pm$ 172 & 5,219  \\
     & Same FFNN     & 752 $\pm$ 171 & 91,523 \\  
     & WANN    & 608 $\pm$ 161 & 245 \\ 
     \hline
   \bottomrule

\end{tabular}
\end{center}
\end{table}

\subsection{Investigating Evolved Neurons}
To gain a better idea of how a neural network with random weights but optimized neuro-centric parameters is solving the task, we plotted all activated neurons of a layer of the champion network (Fig.~\ref{fig:functions}) of the \texttt{CarRacing} environment. The figure shows that while several of the found activations look like the standard form we would expect from a hyperbolic tangent function with a bias, many of the other types of functions also emerged after optimization. We see functions with strong oscillatory behaviors, some in the whole input space, and some only in smaller sections. Other functions have extra peaks and valleys compared to the standard hyperbolic tangent. We hypothesize that this property allows each neuron to respond with more nuance to its input. Additionally, given the same inputs, this collection of neurons responds diversely. As seen in Table \ref{tab:eval}, this evolved diversity of neural computations within a layer allowed the agents to perform well, even though information between layers is projected randomly. For a similar depiction of the activations of the simple, non-recurrent neurons, see Figure~\ref{fig:simple_functions}.

\subsection{Comparison to Weight Agnostic Neural Networks}
An approach similar in spirit to ours is the weight agnostic neural network  (WANN) approach by \citeauthor{gaier2019weight} \cite{gaier2019weight}. As detailed in Section~\ref{related}, in WANNs, only the architecture of the neural network is learned (including choosing an activation function from a predefined set for each neuron) while avoiding weight training. 
While an apples-to-apples comparison is not possible (due to different optimization algorithms), it is nevertheless interesting to see how these two methods compare in terms of performance. Since connections are added to the WANN models during optimization, we cannot directly compare the number of parameters that were optimized in these models to that of the neural units. In Table \ref{tab:eval}, we simply list the final number of synapses in the evolved network structures reported by Gaier and Ha, to give an idea of the network sizes.
The optimized neurons tend to score better in all three environments. These results suggest that it might be easier to optimize customizable neural units for each position in a fully connected network than it is to learn a network structure from scratch. 

In the future, these approaches could be complementary. We imagine that extending the WANN approach with more expressive neurons could allow their evolved neural architectures to become significantly more compact and higher performing.

\section{Discussion and Future Work}
\label{discussion}
In this paper, we introduced an approach to optimize parameterized, stateful neurons. Training these alone yielded neural networks that can control agents in the \texttt{CartPoleSwingUp}, \texttt{CarRacing}, and \texttt{BipedalWalker} environments, even when the weights of the network were never optimized. While optimizing small neural units alone is unlikely to beat state-of-the-art methods on complicated tasks, the neuro-centric optimization alone did nevertheless enable meaningful behavioral changes in the agents. We find these results encouraging, as they pave the way for interesting future studies.

The largest weight-optimized network \textbf{(Same FFNN)} achieved superior scores compared to the neural units in the \texttt{CartPole-\linebreak SwingUp} and \texttt{BipedalWalker} environments. This is not surprising; weight optimization of ANNs now has a long history of success in a plethora of domains. When using random transformations, there is a risk of getting a degraded signal, something that can be compensated for easily by tuning the weights of the transformations. 
Surprisingly, the optimized neural units achieved the best score of the experimental settings in the \texttt{CarRacing} task. The failure of the largest weight-optimized network (\textbf{Same FFNN}) to perform well here might be explained by the relatively low population size compared to the number of parameters being optimized ($128$, and 91,523, respectively). This population size was the same for all experimental settings to ensure that all models are evaluated the same number of times in the environments during optimization. The advantage of having a smaller number of adjustable parameters also came into display in that the larger models could not be optimized by the more powerful CMA-ES method.  

As part of the proposed parameterized neurons, we included a persistent neural state that is fed back to the input of the neuron at the subsequent time step. This endows the network with a memory mechanism. Memory as local neural states is unusual in ANNs, but is much more common in more the biologically inspired Spiking Neural Networks (SNNs) \citep{tavanaei2019deep, pfeiffer2018deep, izhikevich2006polychronization}. Such a neural state is most useful for data with a temporal element, such as agents acting in an environment. It is reasonable to assume that the same approach would have limited use in tasks with unordered data. However, for RL tasks, stateful neurons provide a relatively inexpensive way of allowing the network to have some memory capacity. Setting up more common recurrent neural networks (RNNs), like LSTMs \citep{hochreiter1997long} or GRUs \citep{cho2014properties}, for the tasks used in this paper, would result in the need for many more adjustable parameters than the number of parameters in the neural units optimized here. Combining stateful neurons with more commonplace RNNs could result in interesting memory dynamics on different timescales.

A simpler version of the parameterized neurons with no recurrent state was also tested. Examples of activations of these neurons can be found in \ref{fig:simple_functions}. While a variety of activation curves are displayed, they are all limited to monotonically increase or decrease along the x-axis. However, even the more simple representation of neurons was able to be optimized within a randomly connected network to get relatively high scores - though lower than the recurrent neurons - in all tasks. Especially surprising was the performance in the \texttt{CarRacing} environment, which was close to the score achieved by the recurrent neurons. While memory capacity might be an advantage, it does not seem necessary to perform relatively well in the chosen RL tasks.
Note, that it is straightforward to incorporate different or more information into the neural units. Interesting examples of additional information could be reward information from the previous time step or the average activation value of the layer at the previous time step in order to add some lateral information to the neural activation. 

The ability to improve performance while leaving weights random opens up the possibility for future work of combining neural units like the ones proposed here with the approach of masking weights \cite{frankle2018lottery} mentioned in Section~\ref{related}. An advantage of using masks is that one can train masks for different tasks on the same random network \citep{wortsman2020supermasks}. With pre-trained neural representations, it should be possible to bias the random network to perform actions that are generally useful within a specific task distribution. One could then train masks on top of the untouched weights to perform well on specific tasks, conceivably more efficiently than with generic activation functions.

Another potentially interesting avenue for future work is to combine the optimization of neurons with synaptic plasticity functions \cite{soltoggio2018born}. A lot of work has been done in the area of learning useful Hebbian-like learning rules \citep{chalmers1991evolution, miconi2016learning, mouret2014artificial, floreano2008bio, najarro2020, risi2012unified, soltoggio2008evolutionary, toneli2013,  wang2019evolving, chalvidal2022meta, pedersen2021evolving}. Less work in the field has explored the interaction between learning rules and neural activation in ANNs, despite the fact that most learning rules take neural activations as inputs. It seems likely that more expressive neural units would in turn result in more expressive updates of weights via activity-dependent learning, and thus more powerful plastic neural networks.

Having shown that the proposed neural units can achieve well-performing networks when optimized alone, future experiments will explore the co-evolution of neural and synaptic parameters. It will be interesting to see whether a synergistic effect arises between these two sets of parameters. If both weights and neurons are optimized together, will there then be as much diversity in the resulting set of neural units, or will the need for such diversity decrease?

Despite being initially inspired by biological neural networks, ANNs are still far from their biological counter parts in countless aspects. While the work presented here does not claim to have presented a biologically plausible approach, we do believe that inspiration from biological intelligence still offers great opportunities to explore new variations of ANNs that can ultimately lead to interesting and useful results.

\begin{acks}
 This project was funded by a DFF-Research Project1 grant (9131-00042B). 
\end{acks}

\bibliographystyle{ACM-Reference-Format}
\bibliography{refs}

\end{document}